# Span Identification of Epistemic Stance-Taking in Academic Written English

**Masaki Eguchi** and **Kristopher Kyle**
Learner Corpus Research and Applied Data Science Lab
https://lcr-ads-lab.github.io/LCR-ADS-Home/
Department of Linguistics, University of Oregon
{masakie,kkyle2}@uoregon.edu

## Abstract

Responding to the increasing need for automated writing evaluation (AWE) systems to assess language use beyond lexis and grammar (Burstein et al., 2016), we introduce a new approach to identify rhetorical features of stance in academic English writing. Drawing on the discourse-analytic framework of engagement in the Appraisal analysis (Martin & White, 2005), we manually annotated 4,688 sentences (126,411 tokens) for eight rhetorical stance categories (e.g., PROCLAIM, ATTRIBUTION) and additional discourse elements. We then report an experiment to train machine learning models to identify and categorize the spans of these stance expressions. The best-performing model (RoBERTa + LSTM) achieved macro-averaged F1 of .7208 in the span identification of stance-taking expressions, slightly outperforming the intercoder reliability estimates before adjudication (F1 = .6629).

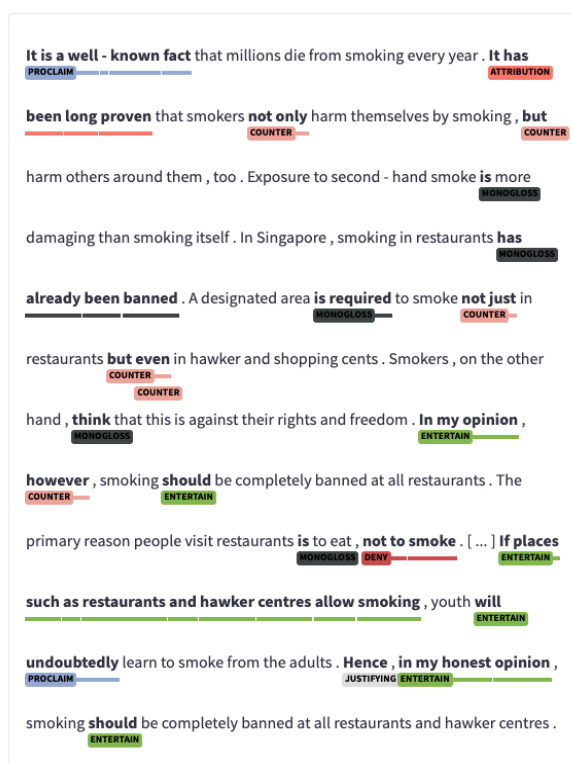

Figure 1: A sample output of the best-performing system reported in this study. The excerpt was taken from the ICNALE corpus (Ishikawa, 2013).

## 1 Introduction

Automated writing evaluation (AWE) systems make it possible to assess students' writings and provide useful feedback efficiently (Shermis & Burstein, 2013). From the language assessment perspective, however, *usefulness* is multifaceted (e.g., Bachman & Palmer, 1996) and, in many parts, depends on what areas of writing ability a given system can measure and give feedback on (Huawei & Aryadoust, 2023). While many AWE systems to date focus on lexical, syntactic, organizational, and topical aspects of students' writing (e.g., Attali, 2007), the construct of writing (i.e., writing skill) is known to be far more complex and includes pragmatic and rhetorical knowledge (Bachman & Palmer, 2010; Sparks et al., 2014). Accordingly, recent studies have included constructs such as discourse moves and steps (e.g., Cotos, 2014), source use and citations (Burstein et al., 2018; Kyle, 2020), and argument structures using Rhetorical Structure Theory (Fiacco et al., 2022). Given the increasing focus on the assessment of the ability to construct effective persuasive texts (Sparks et al., 2014), innovative use of NLP is needed how to assess these social and rhetorical constructs of writing (e.g., Burstein et al., 2016; Carr, 2013; Lu, 2021).

One area that has received relatively little attention in the literature on AWE is the notion of evaluative language or stance-taking (Biber & Finegan, 1988; Hunston & Thompson, 2000; Xie, 2020). In the computational linguistics context, the notion of stance is often discussed in relation to the stance-detection task, where the objective is to categorize whether a text producer is *in favor*, *against,* or *None*, toward a certain topic (e.g., Schiller et al., 2021). However, from the language assessment perspective, researchers are more interested in the rhetorical strategies used to express a nuanced stance instead of the binary classification of positions (Biber, 2006; Biber & Finegan, 1988; Hyland, 2005). In applied linguistics research, evaluative language essentially concerns how writers express their stance on a topic of discussion or express their emotions or feelings on an entity (see Xie, 2020).

This paper reports the development and empirical evaluation of an end-to-end system to identify and categorize epistemic evaluative meanings in academic written discourse (see Figure 1 for illustration). We specifically draw on the discourse-analytic framework of the engagement system in the appraisal analysis (Martin & White, 2005) to create a gold-standard corpus of academic English. We then train an end-to-end span identification systems that can undertake stance analysis under the discourse functional framework. The free online demo of the current span identification system is accessible through Hugging Face Space[1].

## 2 Background

### 2.1 Evaluative language

English for Academic Purposes (EAP) research often investigates evaluative language through corpus-based or discourse-analytic methods (Xie, 2020). Both approaches have both benefits and drawbacks. Qualitative discourse analysis allows researchers to analyze nuanced stance-taking strategies using contextual information; however, this limits the scalability of the analysis and thus cannot be used for large-scale standardized testing situations. Corpus-based approaches (e.g., Bax et al., 2019; Biber, 2006; Yoon, 2017) can overcome the issue of scalability. However, most tools rely extensively on lexical and syntactic features (e.g., dictionary lookups of relevant vocabulary filtered for particular POS tags). Accordingly, these corpus approaches tend to neglect the fact that evaluative language can be poly-functional depending on the surrounding context. For example, very few corpus tools disambiguate whether the verb *suggest* is used to attribute an idea to external sources (*The authors **suggest** that …*) or to hedge the writers' own view (e.g., *We **suggest** that …*). Therefore, a probabilistic approach to identify the function in which the evaluative language is used is necessary to overcome the dilemma faced in the two approaches.

### 2.2 The engagement system

In this study, we draw on the framework of *engagement* in the appraisal analysis (Martin & White, 2005; White, 2003) as a theoretical framework for annotating functional categories of stance-taking expressions. According to Martin and White (2005), engagement concerns "locutions which provide the means for the authorial voice to position itself with respect to, and hence to 'engage' with, the other voices and alternative positions construed as being in play in the current communicative context" (p.94). In this discourse-analytic framework, parts of sentences (or clauses) are classified into different stances writers take. For example, a writer can present his/her idea as if it is a fact (e.g., *The banks **have been** greedy*; Martin & White, 2005). The use of present tense in the example implies that the statement does not recognize potential alternative realities and is thus termed **MONOGLOSS** by Martin & White (2005). Alternatively, a writer can display their awareness of other positions on the topic of discussion, using various heteroglossic strategies. These include, for example, **ATTRIBUTE** (e.g., *I heard on the recent news that the banks have been greedy*), **COUNTER** (e.g., *Although you might disagree*, the banks have been greedy), and **CONCUR** (e.g., *Everyone agrees that the banks are greedy.*), etc. (see a complete list of discourse choices in Section 3.4).

The engagement system has been shown useful in describing nuanced ways in which writers position themselves against possible alternative views, for example, in peer-reviewed academic paper (e.g., Chang & Schleppegrell, 2011; X. Xu & Nesi, 2017), university written assignments (e.g., Lancaster, 2014; Wu, 2007), and second language writing research (e.g., Lam & Crosthwaite, 2018). However, the analysis requires intensive manual coding because of the lack of automated tools that classifies the discourse-semantic category of engagement reliably. This means that in its current state, the engagement system cannot be applied to any large-scale educational applications. To benefit from the theoretical insights of discourse analysis in educational practices, this methodological obstacle needs to be overcome. The current study attempts to fill this gap using a supervised machine-learning approach.

---
[1] https://huggingface.co/spaces/egumasa/engagement-analyzer-demo

## 2.3 Span identification

In this study, the task of identifying the evaluative language of engagement is conceptualized as a span identification task (see Gu et al., 2022; Papay et al., 2020). Span identification is a task of identifying boundaries of expressions in the input text and assigning a label (discourse-semantic one in the current study). Span identification has been used for a range of applications, including entity extraction (Gu et al., 2022), quoted material detection (Pareti, 2016), and toxic word detection (Rao, 2022). Particularly the latter two tasks are directly relevant to the current task because it attempts to identify text segments that may not be easily determined by particular grammatical features (e.g., noun chunks).

Recent span identification architectures (e.g., Gu et al., 2022; Rao, 2022) leverages large encoder-based pre-trained Transformer models (Devlin et al., 2019; Liu et al., 2019). For example, Gu et al. (2022) compared three approaches to formulate span identification tasks—Sequence Tagging, Span Enumeration, and Boundary Prediction. According to Gu et al. (2022), tagging is similar to NER in that each token is predicted under the BIO scheme (e.g., Papay et al., 2020). Span enumeration approaches the task by considering all spans within specified *n* lengths as candidates (as in Lee et al., 2017). Finally, boundary prediction takes a supervised approach to predict the start and end of spans. In the latter two approaches, span representations are created by pooling a set of token embeddings within the candidate spans (e.g., start and end tokens) (see Fu et al., 2021; Gu et al., 2022). Using the RoBERTa-base (Liu et al., 2019) and T5-base encoder (Raffel et al., 2020), Gu et al. (2022) concluded that while the three had relative (dis)advantages, recall-focused tasks may benefit from span enumeration and boundary prediction.

In previous span identification architectures, researchers have often used additional contextualization by adding an additional Bi-LSTM layer on top of the transformer embeddings. However, the results appear mixed depending on the nature of the task and dataset (Gu et al., 2022; Papay et al., 2020). Therefore, a secondary goal of this study is to test whether we observe the benefits of additional contextual information via additional Bi-LSTM when the task does appear to require fine-tuned contextual information due to the discourse oriented nature of the proposed task (see Sections 2.1 and 2.2; see examples of the verb *suggest*).

## 2.4 Contribution of this study

The main contributions of this paper are two-fold. First, we present a new annotation scheme of academic English writing drawing on the discourse-analytic framework of the engagement (Martin & White, 2005) and present annotated dataset using the developed scheme (Section 3). Second, we present a new end-to-end model that can identify and categorize the span of engagement strategies (see Figure 1).

# 3 Engagement Discourse Treebank (EDT)

The EDT currently comprises 4,688 sentences with manually annotated engagement resource spans (126,411 tokens; 11,856 spans), which were sampled from corpora of academic English or closely related genres (see definition of in-domain text below). The version of EDT used to train the machine learning models presented in this paper is accessible at https://github.com/LCR-ADS-Lab/Engagement-Discourse-Treebank. The most recent version of the annotation guideline is accessible through the following GitHub page: https://egumasa.github.io/engagement-annotation-project/.

## 3.1 Definition of in-domain text

When developing a new dataset for an NLP task, it is important to clearly define the domain of texts to sample the annotation data to ensure the correspondence between the gold-standard annotation and the kind of data to make inferences (Ramponi & Plank, 2020). A precise definition of in-domain text is also important from the AWE perspective since the degree of correspondence will influence the degree to which the AWE is able to assess the language use in the Target Language Use domain in language assessment (TLU domain; Bachman & Palmer, 2010). Following these two related concepts, we defined the in-domain text of EDT as academic written English of various genres written by both first- and second-language writers of English.

### 3.2 Source corpora

Annotation data was widely sampled from pre-existing corpora to represent the in-domain texts (see section 3.1 for definition). A major portion of data was sampled from two corpora of university written assignments—the British Academic Written English (Alsop & Nesi, 2009) and the Michigan Corpus of Upper-level Student Papers (Römer & O'Donnell, 2011)—representing first- and second-language writers of English. The remaining portion of data was sampled from a combination of corpora documenting timed essays by second-language writers with various backgrounds and proficiency levels (Blanchard et al., 2013; Ishikawa, 2013; Yannakoudakis et al., 2011). The selection of a wide range of sources, instead of commonly used data sources, such as Wall Street Journal articles, allowed us to represent the characteristics of in-domain texts.

### 3.3 Minimal context approach

During the corpus sampling, we opted for a minimal context window strategy (i.e., three-sentence) to achieve a compromise between the validity of the annotation and any practical considerations (e.g., budget, time constraints, copyrights of source corpora). In an ideal situation, the unit of analysis for annotation should be the entire document, particularly because the object of the annotation is discourse semantics; however, there are arguably advantages and drawbacks to this approach. One advantage of the current three-sentence window approach is that a small dataset (like EDT) can still represent a larger number of writers (hence individual writing styles and stance-taking strategies) compared to using the whole document as a unit of analysis. The coverage of patterns of stance-taking strategies was deemed as important as the annotation of the entire documents, to allow generalization of the machine learning system to different writing styles. A potential drawback of this approach is the reduction of contextual information during annotation; however, using the minimal contexts mitigates this potential issue. This point is taken up in the limitation section, where we offer recommendations and our plans for further research.

### 3.4 Core Engagement Categories

There are eight core engagement categories annotated for EDT. The category definitions and descriptions below were adapted from previous studies (Martin & White, 2005; Wu, 2007; Y. Xu, 2020). The examples are only for illustrative purposes. Note that the ***bold-italics*** in the examples show the spans to be annotated and categorized.

**Monogloss** concerns a statement that does not acknowledge any recognition of potential alternative viewpoints. Such an utterance ignores the dialogic potential in an utterance typically through bare assertions (e.g., *The language you speak **determines** your thoughts*).

**Disclaim-Deny** is an utterance that invokes an alternative position but rejects it directly (e.g., *The language you speak **does not** determine your thoughts*).

**Disclaim-Counter** is an utterance that expresses the idea so as to replace an alternative and thus counter the position which would have been expected (e.g., ***Despite the lack of evidence**, the language you speak determines your thoughts*).

**Proclaim-Concur** concerns an utterance where the writers expect/ assume that their position is easily agreed upon by the putative readers (e.g., ***As we all know**, the language you speak determines your thoughts*).

**Proclaim: Pronounce** is an utterance that shows a strong level of writer's commitment accompanied by explicit emphasis and interpolation, thereby closing down the dialogic space (e.g., ***I contend** that the language you speak determines your thoughts*).

**Proclaim: Endorse** includes utterances that use external sources as warrantable, undeniable, and/or reliable. It shows the writer's alignment with the attributed proposition (e.g., ***The study by Wilson showed** that the language you speak determines your thoughts*).

**Entertain** concerns an utterance that presents the author's position as only one possibility amongst others, thereby opening up dialogic space (e.g., *The language you speak **might** influence your thoughts*).

**Attribute** concerns an utterance where the writer delegates the responsibility of a proposition to a third person (i.e., an external source), thereby opening up the dialogic space (e.g., ***It is often believed*** *that the language you speak determines your thoughts.*).

It is important to reiterate that engagement is a discourse semantic category. This means that while there are some prototypical lexico-grammatical items for each category, the exact function needs to be determined with their co-text in mind, and it is challenging to create an exhaustive list of 'expressions' (see Hunston, 2004).

## 3.5 Supplementary discourse categories

Four supplementary discourse labels were added considering the previous discourse-analytic studies in academic domains (e.g., Hyland, 2005; Nesi, 2021). For other tags annotated, see the annotation guideline.

**Citations** is defined as mentions to an external source(s) in the text in form of in-text or narrative citation (e.g., ***Smith (2000)***; ***(Smith, 2000)***).

**Sources**: Mentions to an external source(s) in the text in form of nominal expressions (e.g., ***A recent paper*** *reports …*).

**Endophoric markers** include a part of the text that refers to information in other parts of its own text (e.g., *X is discussed **in Section 9***).

**Justifying** includes locutions that signal persuasion through justification or substantiation (e.g., *The current discussion is important **because it highlights the key factors of climate change***).

## 4 Annotation Procedure

The annotation team consisted of two primary annotators (undergraduate students; linguistics majors) and the principal investigator (PI) (the first author, who was a Ph.D. candidate in a functional linguistics program and holds a master's degree in Second Language Acquisition and English Language Teaching).

The annotation project comprises the following four steps—annotator training (Section 4.1), iterative consensus building (Section 4.2), independent annotation (Section 4.3), and double-checking and quality assurance (Section 4.4). The annotation comprised two tasks—detecting spans and assigning one functional label for each span.

## 4.1 Annotator Training—orientation and guided practice

Annotator training consisted of an orientation phase followed by guided practice. During the orientation phase, the two annotators were introduced to the basic concepts of SFL and the engagement system (Martin & White, 2005), which were summarized by the PI in the annotation guideline. This included the distinction between *monogloss* and *heterogloss*, the distinction between *contraction* and *expansion*, and distinct strategies (see Sections 2.2). Preliminary topics on the lexico-grammatical analysis were also reviewed as needed (Biber et al., 1999), including the notion of constituency, finite and non-finite clauses, subordinate or embedded clauses, and T-units (Hunt, 1965).

In the guided practice phase, the two annotators went through multiple-stage practice with iterative feedback from the PI. First, they were introduced to the annotation tool, WebAnno version 3.2 (Eckart de Castilho et al., 2016; Yimam et al., 2013). WebAnno was used as the graphical user interface that assists the manual span annotation of engagement resources throughout the annotation project. Second, a sample of 500 sentences was distributed to the annotators. They annotated this training sample independently, which was later checked by the IP for the mastery of the discourse annotation framework. For each annotator, the IP identified the patterns of errors in the training, provided tailored feedback independently, and clarified any concepts in the guideline. This step took the annotation team about 10 weeks (50–100 hours of working time for each annotator).

## 4.2 Iterative consensus building

In adapting the discourse-analytic framework of engagement (Martin & White, 2005), care was taken to update the annotation guidelines to make the descriptions rich and context-specific, as recommended by Fuoli (2018). To this end, the annotation team used the first 200 annotation files for active consensus building. Regular meetings were held to discuss the issues during the annotation of these files after each annotator blindly tagged the data. The resolution strategies were then documented in the annotation guideline. The initial annotation by each annotator was used for the inter-annotator agreement reported in this study.

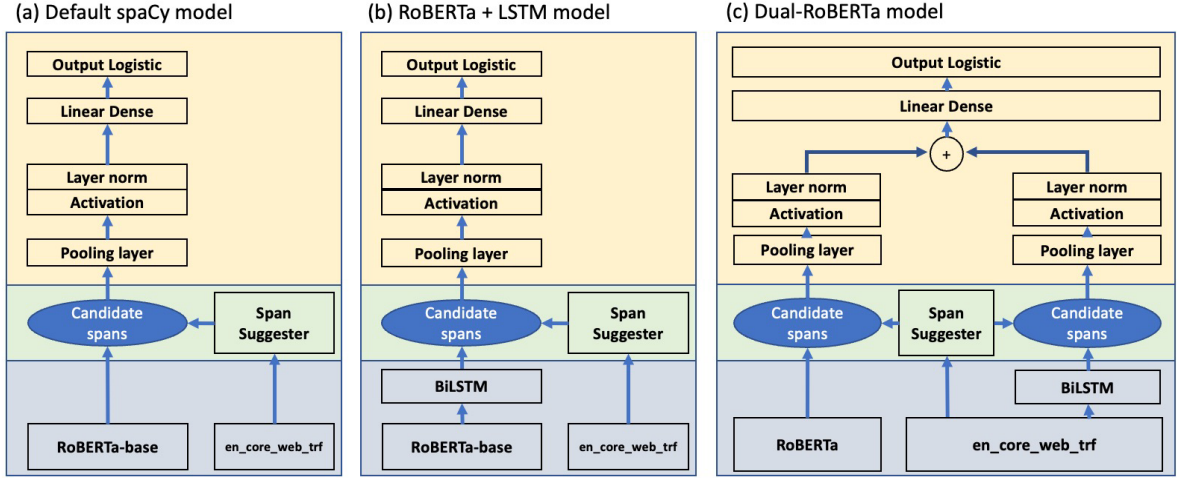

Figure 2: Three architectural variants of the proposed span identification system using the spaCy SpanCat component as the baseline (a). See Sections 5 for details.

### 4.3 Independent annotation phase

Subsequently, the two annotators were assigned to different parts of the corpus. At this point, they were encouraged to document any uncertainties in their annotations and questions in a shared spreadsheet. The annotators were allowed to ask the PI questions about ongoing issues in their annotation, which were mostly addressed in written feedback.

### 4.4 Double-check and quality assurance

Once the data is annotated by either one or two annotators, all the annotation files (both from Sections 4.2 and 4.3) were reviewed by the PI and corrected for any errors and clear deviance from the annotation guideline. After the review of each annotation file, the PI also conducted queries over the entire corpus for any inconsistencies in the spans and the categories. For example, the tag spans for *there is no X* construction (typically DENY) were inconsistently tagged ([*no X*]$_{DENY}$ versus [*there is no X*]$_{DENY}$). These inconsistencies were fixed (*there is* [*no X*]$_{DENY}$ was used), and any ambiguities in the annotation guidelines were fixed for future iterations of the project.

## 5 Model Architectures

The identification and classification of engagement strategies were formulated as a span identification task (e.g., Gu et al., 2022; Lee et al., 2017). Our proposed architectures most closely resemble the span enumeration approach in Gu (2022), where candidate spans are generated greedily (using *n*-grams and dependency subtrees). Figure 2 shows three variants of our neural architecture. We started from the baseline spaCy span categorizer model (Honnibal et al., 2020). We then gradually built the model complexity, guided by previous work in span identification (e.g., Gu et al., 2022; Lee et al., 2017; Papay et al., 2020; Zhu et al., 2021) and our intuitions as linguists. The basic span categorizer pipeline consists of Token embedder, Span Suggester, and Span Categorizer.

### 5.1 Baseline—spaCy Span Categorizer

The first group of ML models uses a single transformer layer as Token Embedder, which is then sent to a pooling layer and logistic regression (see diagram [a] in the Figure 2). This is the default span categorizer implementation provided by spaCy (Honnibal et al., 2020; Schmuhl et al., 2022). In our implementations, we used the off-the-shelf spaCy en_core_web_trf model to predict dependency representations of the input text, which were used to suggest candidate spans along with *n*-grams. For each candidate span, span representation is created by taking the RoBERTa-base embeddings and applying several pooling operations. The pooled span representation is sent to the non-linear activation function and subsequently to the logistic layer for prediction. In this architecture, the RoBERTa embeddings were fine-tuned to learn task-specific weights while the weights from the en_core_web_trf model were fixed.

## 5.2 RoBERTa + Bi-LSTM model

Although Transformer models can provide contextually aware token representations (Clark et al., 2019), it was hypothesized that additional sequential information would be beneficial for classifying engagement strategies that are interpreted by discourse analysts with co-textual information in mind, such as ATTRIBUTION. To allow the model to learn this additional contextual information, we added a single-layer Bidirectional Long-Short Term Memory (Bi-LSTM; Hochreiter & Schmidhuber, 1997; Schuster & Paliwal, 1997) architecture on top of the RoBERTa token-level embeddings, before they are sent to the span pooling layer. Such architecture has often been implemented in previous span identification architectures (Gu et al., 2022; Lee et al., 2017; Papay et al., 2020; Zhu et al., 2021). For the purpose of the current study, we used one-layer Bi-LSTM with 200 hidden dimensions following the previous study (Gu et al., 2022).

## 5.3 Dual-RoBERTa model

The third architecture used two sets of transformer embeddings side-by-side, concatenated before the final output layer for prediction (see Architecture (c) in Figure 2). This model architecture was inspired by recent ensemble approaches to span identification pipelines (e.g., Rao, 2022). The intuition behind the dual-Transformer architecture was that the two Transformer models would offer complementary information to categorize the span labels, particularly because the second Transformer layer from the spaCy en_core_web_trf model was already fine-tuned for multitask learning objectives (e.g., POS tagging, Dependency parsing, Named Entity Recognition) on the Ontonote 5.0 corpus (Weischedel et al., 2013). Note that the RoBERTa weights from the en_core_web_trf was fixed in order to avoid forgetting of the important information for the dependency parsing.

## 5.4 Domain adaptation of RoBERTa

Since the version of EDT used for training was still relatively small, adaptive pre-trainings were conducted on the RoBERTa-base model (Liu et al., 2019) using the checkpoint available through Hugging Face library (Wolf et al., 2020) in hope to counteract potential mismatches between the RoBERTa embedding and the characteristics of in-domain texts (Han & Eisenstein, 2019; Ramponi & Plank, 2020). To this end, four domain adapted RoBERTa-base models were created. The five versions of RoBERTa (including the original) were set as hyperparameter in the following experiment (see Appendix B).

## 6 Methods

We implemented the three architectures through spaCy version 3.4 (Honnibal et al., 2020). All models were trained on a quad Nvidia Tesla K80 GPU with 12GB RAM. All models were optimized with Adam Optimizer.

### 6.1 Data preparation

Table 1 summarizes the number of tags by category in the dataset used for this experiment. Two pairs of tags (Concur and Pronounce; Endorse and Attribute) were collapsed as PROCLAIM and ATTRIBUTION, respectively, to obtain enough number of instances in the dev and test sets. According to the engagement system (Martin & White, 2005), Concur and Pronounce are subtypes of PROCLAIM strategy along with ENDORSE, while ENDORSE was categorized under ATTRIBUTION with Attribute in this study due to its primary function of such (Sections 2.2 and 3.4). We then created five sets of 80/10/10 splits for 5-fold cross-validations (CV). The tag counts in the 5-fold datasets can be found in Appendix A. Due to the imbalances in labels, we oversampled minority cases in each data split (after splitting them into training sets to avoid data leaks). The oversampling approach (e.g., Wang & Wang, 2022) was used because there is no existing model to create synthetic examples for this new type of NLP task.

| Category | Tag counts |
| --- | --- |
| ATTRIBUTION | 1247 |
| COUNTER | 1046 |
| DENY | 887 |
| ENTERTAIN | 2837 |
| MONOGLOSS | 2742 |
| PROCLAIM | 445 |
| CITATION | 618 |
| ENDOPHORIC | 213 |
| JUSTIFYING | 966 |
| SOURCES | 855 |

Table 1: The number of tags by category in the entire EDT. ATTRIBUTION subsumes ATTRIBUTE and ENDORSE; PROCLAIM subsumes CONCUR and PRONOUNCE in the original tags.

|  | Human annotation baselines | | End-to-end models trained on adjudicated data | | | | | |
| --- | --- | --- | --- | --- | --- | --- | --- | --- |
|  | *Read & Carroll (2012)* | *Our annotator agreement* | spaCy default | | RoBERTa+LSTM | | Dual-RoBERTa | |
| *Category* |  |  | *M* | *Min* | *M* | *Min* | *M* | *Min* |
| ATTRIBUTION | .379 | .5943 | .6969 | .6553 | **.7127** | .6761 | .6911 | .6149 |
| COUNTER | .603 | .8511 | .8521 | .7394 | .8636 | .7781 | **.8774** | .8567 |
| DENY | .451 | .8621 | .8570 | .8257 | .8800 | .8579 | **.8815** | .8522 |
| ENTERTAIN | .459 | .8278 | .8413 | .7917 | **.8360** | .7755 | .8340 | .7903 |
| MONOGLOSS | n/a | .8092 | **.8017** | .7476 | .7864 | .7568 | .7890 | .7314 |
| PROCLAIM | .336 | .4038 | .6685 | .6127 | .6906 | .6203 | **.7027** | .6197 |
| CITATION | n/a | .9497 | .9047 | .8875 | .9185 | .8953 | **.9193** | .9015 |
| ENDOPHORIC | n/a | .6071 | .7236 | .6000 | .7254 | .6316 | **.7418** | .6919 |
| JUSTIFYING | n/a | .8203 | .8131 | .7766 | **.8167** | .7404 | .8081 | .7608 |
| SOURCES | n/a | .5663 | .6961 | .6585 | **.6985** | .6318 | .6844 | .5887 |
| Accuracy |  | .7146 | .7015 | .6885 | **.7095** | .6960 | .7054 | .6922 |
| macro avg F1 |  | .6629 | .7141 | .6942 | .7208 | .7105 | **.7209** | .7108 |
| weighted avg F1 |  | .7208 | .7183 | .7094 | **.7283** | .7105 | .7196 | .6903 |
| Cohen's Kappa |  | .6686 | .6647 | .6509 | **.6738** | .6596 | .6694 | .6549 |
| MMC |  | .6691 | .6663 | .6534 | **.6755** | .6611 | .6710 | .6554 |

Table 2: F1 scores based on 5-Fold CV. Our intercoder agreement is presented side by side with the result reported in Read and Carroll (2012), who annotated the entire Appraisal framework. Due to the adaptations of the original Martin and White (2005) in our study (see Section 3.4) some of the tags lacks direct comparisons. Three neural architectures are compared using the mean and minimum F1 scores based on the 5-Fold CV. MCC = Matthews Correlation Coefficient. Averaged F1 scores were calculated including empty tags.

## 6.2 Hyperparameter tuning and 5-fold cross-validation

We randomly searched the optimal combination of hyperparameters for each of the three architectures and tested the stabilities of the top three settings from each architecture (see Appendix B for hyperparameters). A total of 205 models were trained across the three architectures. Subsequently, eight top-performing hyperparameter settings were chosen, and we then conducted 5-fold cross-validation for each. We report the result of the best 5-fold CV result for each architecture.

## 6.3 Evaluation metrics

Considering the imbalanced data, the models were evaluated using Matthews Correlation Coefficient and Cohen's Kappa on the end-to-end span categorization results. Because our span suggester used span enumeration approach (Gu et al., 2022) and was constant across the models, they were not compared. Note that preliminary experiments showed that the current span suggester (See Appendix B for hyperparameter settings) achieved recall of 97–99% on Development and Test sets.

## 7 Results

Table 2 reports on the inter-annotator reliability and the results of the 5-fold CV.

### 7.1 Inter-annotator agreement

A subset of blind annotation (35,640 tokens; 1,373 sentences; 3,732 unique spans) was used to compute the inter-annotator agreement between the two annotators. The results indicated that the agreement was moderate (Cohen's Kappa = .6686; Matthews Correlation Coefficient = .6691). Comparing the by-tag F1 scores against those by Read and Carroll (2012), our annotator agreement was substantially higher. However, the results also indicate there were some areas of struggle by human annotators (e.g., ATTRIBUTION, PROCLAIM).

### 7.2 Result of the end-to-end models

Overall, the end-to-end models, which were trained on a fully reviewed/adjudicated dataset, tended to outperform the benchmarks of inter-annotator agreement. The gains were substantial in several categories that were challenging for our annotators, including ATTRIBUTION, PROCLAIM, and SOURCES.

### 7.3 Comparison among three architectures

The result of the 5-fold CV (Table 2) indicated that the RoBERTa + LSTM architecture performed best among the three architectures (Cohen's Kappa = .6738; Matthews Correlation Coefficient = .6755). This was followed by the Dual-RoBERTa Model (Cohen's Kappa = .6694; Matthews Correlation Coefficient = .6710). It appears that RoBERTa + LSTM model and Dual-RoBERTa model may complement their strengths and weaknesses.

## 8 Discussion

The results of the 5-fold CVs indicated that the proposed architectures performed as well as (or even outperformed) the inter-annotator agreement baseline set for the study. The results also suggested that our RoBERTa + LSTM and Dual-RoBERTa models tended to perform better than the spaCy default spancat model (Honnibal et al., 2020; Schmuhl et al., 2022).

It is noteworthy that the additional Bi-LSTM layer appeared to enhance the stability of the model. Although the use of a Bi-LSTM layer on top of a Transformer encoder is not uncommon in span identification tasks, its reported benefits have been mixed (Gu et al., 2022; Papay et al., 2020; Zhu et al., 2021). The gain in this study can be explained in two ways—additional sequential information and dimensional reduction. In a simple explanation, the architecture benefited from the additional sequence information provided by Bi-LSTM. At least one previous study (Gu et al., 2022) reported similar gains in additional LSTM layer, particularly when the span suggestion components were similar to the current greedy approach. Thus, it could be that the additional LSTM helped to refine the embedding for this particular span enumeration architecture (Gu et al., 2022; Lee et al., 2017). In addition to this explanation, it is also possible that LSTM worked as a dimension reducer (while maintaining direct sequential information). Future research may clarify the potential reasons for this stability in the span identification task (which is out of the scope of the current study).

Apart from the machine learning experiment, our inter-annotator agreement showed that the span annotation of engagement resource may be a challenging task, particularly for undergraduate annotators (linguistics majors) who were trained over 10 weeks. However, our annotator agreement substantially improved upon the previously published benchmark by Read and Carroll (2012). The moderate reliability in this study may provide further evidence to Fuoli's (2018) claim regarding the lack of explicit guidelines and methodological discussions pertaining to the identification of engagement resources in discourse samples. Thus, it is hoped that the present annotation guideline may serve as a resource to guide future methodological improvement in discourse annotation of engagement resource analysis (see Fuoli, 2018; Read & Carroll, 2012).

## 9 Conclusion

In this paper, we reported a new approach to identifying stance-taking expressions in English texts in academic domains. Specifically, we introduced a new human-annotated corpus of academic English that draws on a discourse-analytic framework of the engagement system from the Appraisal framework (Martin & White, 2005). We also reported an end-to-end system that can conduct automated span identification of stance-taking strategies based on the engagement framework. The experimental result indicates that the system can outperform inter-annotator reliability estimates by a 5–6% gain in the macro-averaged F1 score. The finding, although preliminary, opens a new avenue for feature engineering for the next-generation AWE systems (Burstein et al., 2016), expanding the constructs measured by the AWE engines. A follow-up study by the first author shows that the engagement features can explain the writing scores above and beyond the existing linguistic features at the levels of lexis, grammar, and cohesion (Eguchi, 2023). While end-to-end score prediction models may be used to obtain accurate score predictions, the features introduced in this paper may be used in conjunction with such end-to-end scoring engines to maintain the explainability and interpretability of the scores. The visualization of stance-taking features (see Figure 1 and demo) can also be presented to learners to highlight the patterns of stance-taking in both model and student essays.

### 9.1 Limitations and Future Directions

For future work, we plan to update the annotated corpus and the scope of annotation to paragraphs and/or whole documents. In this study, we opted for the minimal context approach for practical reasons, such as budget, time constraints, and copy rights of the source corpora. The minimal context approach allowed the annotation sample to represent as many writers as possible for better generalization with relatively small sample sizes. However, future research should use longer units of analysis to enhance the quality of manual annotation. Despite this limitation, the results of the current study indicated that the current approach is a promising direction for further research on automated analyses of rhetorical features.

## Acknowledgments

We thank three anonymous reviewers for their insightful comments, which improved the quality of the current manuscript. We express our sincere gratitude to two undergraduate annotators, Aaron Miller and Ryan Walker, for their meticulous work on the discourse annotation. This work was supported by the following grants/awards: the Duolingo English Test's Doctoral Dissertation Award 2022, the International Research Foundation for English Language Education (TIRF) Doctoral Dissertation Grant 2022, the National Federation of Modern Language Teachers Association and the Modern Language Journal (NFMLTA-MLJ) Dissertation Writing Support Grant 2022, the Graduate Student Research Award at the Linguistics Department, University of Oregon, and Dr. Kristopher Kyle's institutional research funds.

## A  Tag counts in each training fold (Before oversampling)

| | Fold 1 | | | Fold 2 | | | Fold 3 | | | Fold 4 | | | Fold 5 | | |
|---|---|---|---|---|---|---|---|---|---|---|---|---|---|---|---|
| | Train | Dev | Test | Train | Dev | Test | Train | Dev | Test | Train | Dev | Test | Train | Dev | Test |
| ATTRIBUTION: | 1028 | 118 | 101 | 995 | 136 | 116 | 993 | 115 | 139 | 987 | 138 | 122 | 985 | 129 | 133 |
| COUNTER: | 879 | 88 | 79 | 818 | 112 | 116 | 839 | 105 | 102 | 848 | 103 | 95 | 800 | 127 | 119 |
| DENY: | 705 | 83 | 99 | 712 | 86 | 89 | 712 | 84 | 91 | 720 | 94 | 73 | 699 | 88 | 100 |
| ENTERTAIN: | 2306 | 254 | 277 | 2262 | 314 | 261 | 2246 | 280 | 311 | 2310 | 248 | 279 | 2224 | 334 | 279 |
| MONOGLOSS: | 2223 | 256 | 263 | 2179 | 264 | 299 | 2184 | 289 | 269 | 2157 | 296 | 289 | 2225 | 273 | 244 |
| PROCLAIM: | 343 | 47 | 55 | 353 | 54 | 38 | 358 | 33 | 54 | 375 | 33 | 37 | 351 | 41 | 53 |
| ENDOPHORIC: | 168 | 29 | 16 | 175 | 18 | 20 | 176 | 12 | 25 | 168 | 30 | 15 | 165 | 21 | 27 |
| JUSTIFYING: | 788 | 84 | 94 | 783 | 93 | 90 | 750 | 118 | 98 | 795 | 79 | 92 | 748 | 108 | 110 |
| CITATION: | 517 | 56 | 45 | 497 | 56 | 65 | 504 | 48 | 66 | 479 | 71 | 68 | 475 | 60 | 83 |
| SOURCES: | 700 | 76 | 79 | 686 | 85 | 84 | 682 | 79 | 94 | 660 | 111 | 84 | 692 | 87 | 76 |
| sum | 9657 | 1091 | 1108 | 9460 | 1218 | 1178 | 9444 | 1163 | 1249 | 9499 | 1203 | 1154 | 9364 | 1268 | 1224 |

## B  Hyperparameters for random search

| Category | Hyperparameter | Possible values (Parameter range or choice) | Selection |
|---|---|---|---|
| Entire model | Model Architecture | Single-Transformer; Single-Transformer+ LSTM; Dual-Transformer + single-LSTM | discrete |
| Token Embedder | Pre-Trained language model | roberta-base; egumasa/roberta-base-academic3; egumasa/roberta-base-university-writing2; egumasa/roberta-base-research-papers | discrete |
| Span Categorizer | FFN (Activation function) | Maxout (default selection by spaCy); Mish; Mish with two separate FFNs | discrete |
| Span Categorizer | FFN (hidden unit sizes) | [128, 256, 384] | discrete |
| Span Categorizer | FFN (dropout rates) | [0, 0.2, 0.3, 0.4] | discrete |
| Span Categorizer | FFN (layer depths) | [1, 2] | discrete |
| Training | Maximum learning rate (alpha) | 6e-5 – 2e-5 | uniform distribution |
| Training | System seed during training | [0, 808, 1993, 1234, 2023] | discrete |
| Training | Gradient accumulation steps | [4, 8] | discrete |
| Span Suggester | Max n-gram lengths | 12 words | fixed |
| Training | Optimizer | Adam with weight decay | fixed |
| Training | Learning rate schedule | linear decay with warm-up steps | fixed |
| Training | Warm-up steps | 1,000 | fixed |
| Training | Maximum training step | 20,000 | fixed |
| Training | Steps before early stop | 3,000 | fixed |
| Training | mini-batch size | defined by number of words | fixed |
| Training | minimal start batch size | [300, 500, 900] | discrete |
| Training | Maximum batch size | 1,000 words | fixed |